\documentclass[conference]{IEEEtran}
\IEEEoverridecommandlockouts
\usepackage{cite}
\usepackage{amsmath,amssymb,amsfonts}
\usepackage{algorithmic}
\usepackage{graphicx}
\usepackage{textcomp}
\usepackage{xcolor}

\usepackage{balance}  % for useful for balancing the last columns
\usepackage{booktabs} % for pretty table rules
\usepackage{ccicons}  % for Creative Commons citation icons
\usepackage{ragged2e} % for tighter hyphenation
\usepackage{subfigure}
\usepackage{multirow}

\def\BibTeX{{\rm B\kern-.05em{\sc i\kern-.025em b}\kern-.08em
    T\kern-.1667em\lower.7ex\hbox{E}\kern-.125emX}}
\begin{document}

\title{Inferring User Facial Affect in Work-like Settings\\
\thanks{This work is supported by the EU's Horizon 2020 Research and Innovation programme project WorkingAge under grant agreement No. 826232.}
}

\author{\IEEEauthorblockN{ Chaudhary Muhammad Aqdus Ilyas, Siyang Song, and Hatice Gunes}
\IEEEauthorblockA{\textit{Affective Intelligence \& Robotics Lab} \\ \textit{Department of Computer Science  \& Technology} \\
\textit{University of Cambridge, UK}\\
cmai2@cam.ac.uk, ss2796@cam.ac.uk, hatice.gunes@cl.cam.ac.uk}

%\and
%\IEEEauthorblockN{}
%\IEEEauthorblockA{\textit{Affective Intelligence \& Robotics Lab} \\ \textit{Dep. of Computer Science  \& Technology} \\
%\textit{University of Cambridge, UK}\\
%ss2796@cam.ac.uk}
%\and
%\IEEEauthorblockN{Hatice Gunes}
%\IEEEauthorblockA{\textit{Affective Intelligence \& Robotics Lab} \\ \textit{Dep. of Computer Science  \& Technology} \\
%\textit{University of Cambridge, UK}\\
%hatice.gunes@cl.cam.ac.uk}
% \and
% \and
% \and
}

\maketitle

\begin{abstract}

% There is growing interest in the implementation of tools to monitor affective states in naturalistic working environments.  In this article we explored the development of affect analysis system in work-like environment considering the challenges of cognitive load, worker's performance and physiological activities. For this purpose, we simulated the work-like environment and collected the facial data from twelve participants performing work-like tasks. Our results showed that people in working environments display subtle emotional states than in non working environment.  This work is of great importance for affect recognition in terms of valence and arousal: with mapping of predictions to a user being in a positive, neutral or negative as well as in active, neutral or inactive affective states.

Unlike the six basic emotions of happiness, sadness, fear, anger, disgust and surprise, modelling and predicting dimensional affect in terms of valence (positivity - negativity) and arousal (intensity) has proven to be more flexible, applicable and useful for naturalistic and real-world settings. 
%to  reliable in the identification of users' emotional and mental states such as happy, sad, active, inactive, feeling positive or negative as well as their well-beings, which can further contribute to improve workers' efficiency and safety in working environments. 
In this paper, we aim to infer user facial affect  when the user is engaged in multiple work-like tasks under varying difficulty levels (baseline, easy, hard and stressful conditions),  including (i) an office-like setting where they undertake a task that is less physically demanding but requires greater mental strain; (ii) an assembly-line-like  setting that requires the usage of fine  motor  skills; and (iii) an office-like setting representing teleworking and teleconferencing.
%To achieve this, we design a study protocol employing 12 subjects from the University, simulating work-like environment and executing work-like tasks under various stress conditions. 
In line with this aim, we first design a study with different conditions and gather multimodal data from 12 subjects. We then perform several experiments with various machine learning models and find that: (i) the display and prediction of facial affect vary from non-working to working settings; (ii) prediction capability can be boosted by using datasets captured in work-like context; and (iii) segment-level (spectral representation) information is crucial in improving the facial affect prediction.        

%to investigate:  (i)
%Does facial affect (valence  and  arousal) predicted by the machine learning models vary significantly in work-like settings as compared to non-working conditions? (ii) How does  segment-level information influence the accuracy of the facial affect predictions? and (ii) How   well   the   predictions  generated by the model  in  terms  of  valence  and  arousal,  match  the self-reported  affect  reported using  the  Self-Assessment  Manikin (SAM)?
%
%dimensional affective states (valence and arousal) assessment in work-like settings;  ii) impact of work-like settings (use of working-environment-context-aware (WECARE-DB) database) in the assessment of affective states; and iii) impact of spectral representation in the prediction of affective states. 

% create the working-environment-context-aware network for emotion recognition (WECARE-NET) by focusing on the assessment of facial affect when humans perform work-like tasks. To achieve this, we designed a study protocol simulating work-like settings that was undertaken across four sites located in the UK, Germany and Italy employing 44 participants performing various work-like tasks.We perform several experiments to investigate: (i) dimensional affective states  (valence and arousal), (ii) the impact of context and background and (iii) the effectiveness of temporal segment length in the assessment of valence and arousal. 
% Our results show that: (i) context information boosted the prediction accuracy and (ii) too short and too long segment length reduces the prediction performance.

\end{abstract}


\section{Introduction}

%This work aims to infer the workers' affect in different working environments through the usage of non-obtrusive sensors such as cameras. To validate the findings of the scientific investigation, these emotional states are compared to different working environments.  
There are different ways of modelling and analysing human affect. The theory of six basic emotions (happiness, sadness, surprise, fear, anger and disgust) has been a good simplification but in the real-world, e.g., where people live and work, people do not display emotions in an exaggerated manner that can be categorized into these six categories. The dimensional perspective has been proposed as a viable alternative for non-acted emotions. It suggests that emotions are responses to environmental stimuli that vary along three key dimensions - valence/pleasure, arousal, and dominance/control. This approach has now been widely adopted by the affective computing community – see \cite{gunes2010automatic}, \cite{gunes2013categorical} for extensive reviews. 

Previous research studies explored facial affect recognition through hand-crafted features \cite{mohammadi2014pca, Uddin2017Facial} and deep neural networks \cite{ ilyas2021deep, ilyas2018rehabilitation, ilyas2020deepface,chen2018softmax}. However, all these studies are conducted on facial datasets recorded in the lab (controlled) conditions or in-the-wild (data scrapped from the internet) settings (please see \cite{mollahosseini2017affectnet} for details). 
%
%There has been a vast research in the field of emotional states analysis across various disciplines such as psychology, neuroscience, sociology and computer science\cite{sariyanidi2014automatic}. 
Systems created for ambient assistive living  environments \cite{gunes2010automatic, calvaresi2017exploring,  maskeliunas2019review} aim to be able to perform both automatic affect analysis and responding.
%Advanced intelligent systems in smart and ambient assistive  living (AAL) environments \cite{gunes2010automatic,  calvaresi2017exploring, maskeliunas2019review} are equipped with systems capable of Automatic Affect Analysis and Response (AAFAR) accordingly.
Ambient assistive living relies on the usage of  information and communication technology (ICT) to aid in person's every day living and working environment to keep them healthier and  active longer, and enable them to live independently as they age. Thus, ambient assistive living  aims to facilitate health workers, nurses, doctors, factory workers, drivers, pilots, teachers as well as various industries via sensing, assessment and intervention. The system is intended to determine the physical, emotional and mental strain and respond and adapt as and when needed, for instance, a car equipped with a drowsiness detection system can inform the driver to be attentive and can suggest them to take a little break to avoid accidents \cite{Reddy_2017_CVPR_Workshops, Deng_Driver_2019}. 

%Workforce thoughts, feelings, and actions are affected by the organization and environment in which they work. Similarly, workers feelings, thoughts and actions impact the organization \cite{brief2002organizational}. A study performed by Porras and Hoffer \cite{porras1986common} in 1986 investigated the workers behavior in the positive organizational outcome from 42 top organizations. He identified that workstyle behaviors (communication, feelings, gestures, stress and safety of the workers) are key identifiers associated with better performance. Similarly,  several studies have identified the strong relationship between job experience and mood or affect at work \cite{basch2000affective, wegge2006test } and workers mood in a working settings \cite{judge2004affect,casper2019power, totterdell2003emotion}. Affect symmetry  for example 

Research shows that employees' moods, emotions, and dispositional affect influence critical organizational outcomes such as job performance, decision making, creativity, turnover, prosocial behaviour, teamwork, negotiation, and leadership \cite{barsade2007does}. Therefore, analysing and understanding the affect of the employees in an organisational setting is crucial in shaping organizational behaviours and decisions. To understand and evaluate the emotional and affective states in working environment it is necessary to gather data from the actual working environment or work-like settings as emotional and affective states in these environments vary from other settings due to the specific physical and mental workload, and physiological activity of the worker \cite{giorgi2021wearable}. 
More importantly, workers' affect in working environments relate to their performance, well-being, risk perception and assessment, and can be even used for quality control \cite{borghini2020stress, borghini2014measuring}. 

Therefore, in this paper, we aim to train and evaluate machine learning models to infer user facial affect when the user is engaged in multiple work-like tasks under varying difficulty levels. More specifically, we explore (i) an office-like setting where they undertake a task that is less physically demanding but requires greater mental strain; (ii) an assembly-line  setting that requires the usage of fine  motor  skills; and (iii) an office-like setting representing teleworking and teleconferencing.

The rest of this paper is organised as follows: Section \ref{sec:study} describes the study protocol developed to acquire data, and tasks performed to simulate work-like settings. Section \ref{sec:method} presents the methodology and section \ref{sec:results_discussion} analyses the results of the experiments conducted and concludes the paper.

\section{The Study}
\label{sec:study}

\subsection{Study Aims}

To investigate the people's affective and emotional states in work-like settings, we designed a study and conducted different experiments that simulate the working environment conditions and challenges such as varying mental workload, varying physical load and varying stress levels. With this, we aim to investigate the following research questions: 

%% Siyang modified

% In this paper, we present the first study that specifically investigate the human affective states in working environments. In particular, we propose a novel protocol to collect human audio-visual and psychological data under multiple simulated work-like-setting tasks. The goal of the work-like-setting tasks is to stimulate participants in various ways to acquire naturalistic facial affect data that correspond to their mental workload, stress, and affective state by asking them undertaking various pre-defined tasks. Based on this protocol, we built a novel database called Context-Aware-Working-Environment (CAWE). We then propose a framework called Working-Environment-Context-Aware Emotion Recognition Network (WECARE-NET), i.e. a linear combination of Convolutional Neural Networks (CNNs) and Long-short-Term-Memory Networks (LSTMs), to investigate the influence of spatio-temporal contextual information on face-based dimensional affect recognition. In summary, we aim to investigate the following research questions: 

\begin{itemize}
\item RQ1: Does facial affect (valence  and  arousal) predicted by the machine learning models vary significantly in work-like settings as compared to non-working conditions?
\item RQ2: How does  segment-level information as compared to frame-level information influence the accuracy of the facial affect predictions? 
\item RQ3: How   well   the   predictions  generated by the models in  terms  of  valence  and  arousal  match  the self-reported  affect  reported using  the  Self-Assessment  Manikin (SAM) - i.e., to what extent are the self-reported labels and system predicted labels agree? ?
\end{itemize} 
%    \item RQ1: Are the affective states (valence and arousal) and emotional states (happy, sad, angry etc.) obtained by trained models (automatically) vary significantly in work-like settings as compared to non-working conditions.  
%    \item RQ2: How much does segment-level information influence in the accurate prediction of affective states.  
%    \item RQ3: Assess how well the predictions our system generates in terms of valence and arousal, match the self-reported affect using the Self-Assessment Manikin (SAM) – i.e., to what extent are the self-reported labels and system predicted labels agree? 
%\vspace{-4mm}

\begin{figure}
	\centering
	\includegraphics[width=6cm]{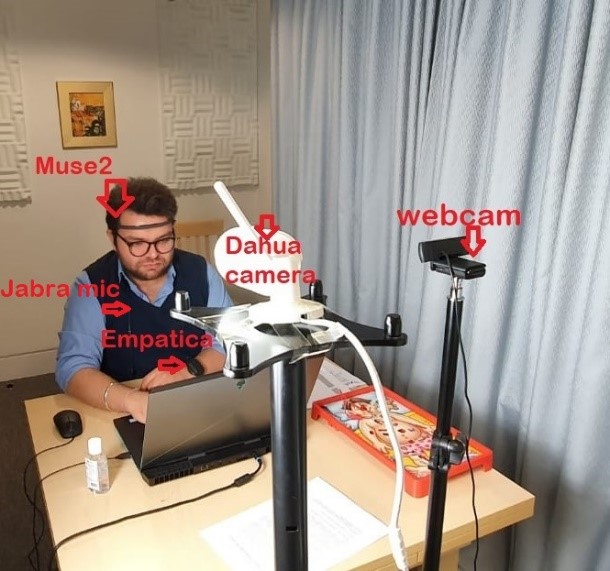}
	\caption{Illustration of the recording and data collection setting.} \vspace{-2mm}
    \label{fig:hardware}
\end{figure}\vspace{-1mm}

\subsection{Recording Setup} 

\noindent The recording setup is illustrated in Fig. \ref{fig:hardware}. A participant is asked to sit at a table, where a laptop displaying slides with instructions is provided to guide the participant through the required tasks. Meanwhile, two cameras (a Logit web camera and a Dahua IP camera) are placed in front of the participant and a GoPro camera is placed on the table. In addition, the participant is also asked to wear three sensors, i.e., a Jabra microphone around their neck that records the participant's voice, an Empatica wristband and a Muse sensor that record psychological signals. As a result, the dataset contains multi-modal recordings for each participant, including audio, video, and a set of psychological signals. In this paper, only the Dahua IPC-HFW1320S-W Camera recordings are considered for facial affect analysis . %The camera needs to be mounted where the face of the user can be frontally captured (e.g., on the screen (or the desk/wall) in front of the user for Office and Teleworking environments).

\subsection{Participants} 

\noindent This study was approved by the University of Cambridge’s Department of Computer Science and Technology Ethics Committee. Following an explanation of the study and informed consent from each participant, the experiment was carried out in accordance with the principles outlined in ethical approval. Additional COVID related measures were also put in place prior to undertaking the study. Twelve participants were recruited from the University of Cambridge (5 male and 7 female  from $9$ countries) with an average age of 28.25 (max=41 and min=22). All participants were proficient in English.

%% Siyang added

\subsection {Data Collection Protocol}

% \noindent A standard protocol was designed to assess affect in work-like contexts. The experiments to stimulate working conditions included three tasks: the N-back task, the Operations Game task (they were looking down during this task),
% %
% %the Doctor Game task, 
% and the Webcall task. Each task contains four different sub-tasks performed in varying challenging conditions namely baseline, easy, hard and hard-under-stress condition. 

\begin{figure*}
\centering
\includegraphics[width=17cm]{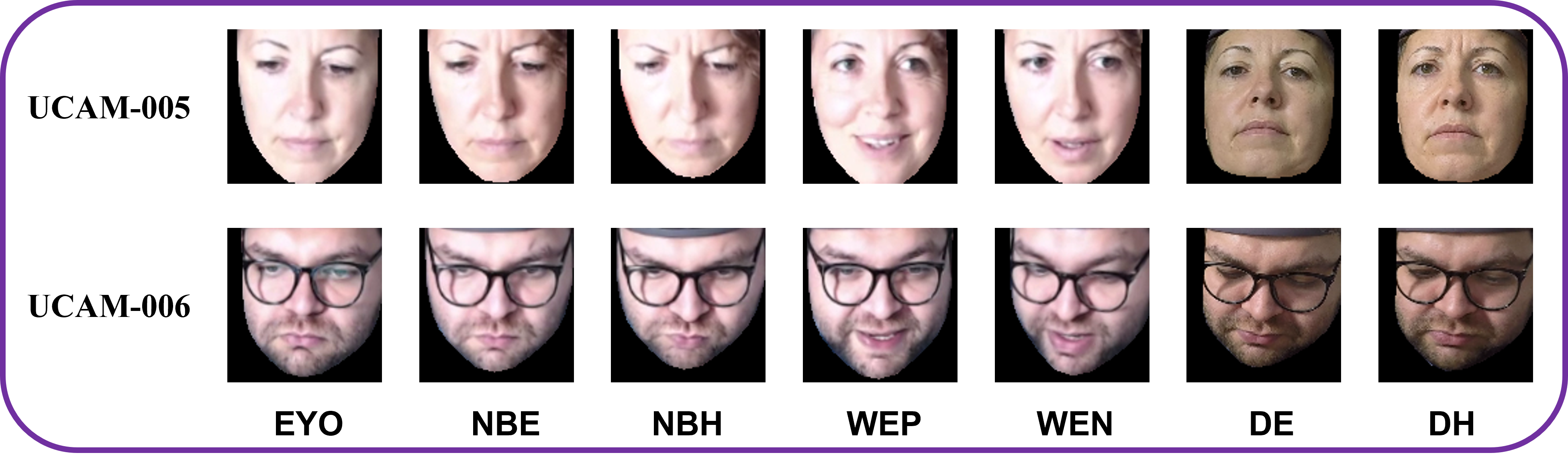}
\caption{Example facial expressions displayed by two participants (005 and 006) and triggered by different tasks, where facial expressions during the webcall tasks are generally more expressive.}\vspace{-1mm}
\label{fig:Facial displays}
\end{figure*}

\noindent To simulate various working conditions,a standard protocol was designed to assess affect in work-like contexts. The experiments to stimulate working conditions included three tasks: the N-back task, the Operations Game task (they were looking down during this task), and the Webcall task. In N-back and Operations Game tasks, participants were asked to conduct different sub-tasks performed in varying challenging conditions namely baseline, easy (conducted twice), hard (conducted twice) and hard-under-stress conditions. For the Webcall task, there are three sub-tasks including baseline, conversation of a happy memory and conversation of a negative memory. The example facial displays triggered by different tasks are visualized in Fig. \ref{fig:Facial displays}.

\textbf{The N-back task:} The activities in this task represent the office-like settings (less physical but with greater mental strain) and test the worker's memory load for a reasonable approximation of work load. In this task, a series of letters are presented on a computer screen and the participant is requested to press the button when the letter on the screen match the letter that appeared in previous $ n $ stages. The task's complexity can be modified by increasing the value of $ n $, challenging participants to memorize more letters in order. In this study, the tasks are categorized into Baseline and three conditions, Easy, Hard and Hard-under-stress. Under all conditions, 21 Uppercase letters (33$\%$ target letters) were exhibited for 500 milliseconds with random inter-stimulus interval of 500 to 3000 milliseconds.  

\begin{itemize}

    \item Baseline (NBB): Participant observes the screen and the letters without pressing any button. 
    \item Easy-0-back (NBE): The participant indicates when the current stimulus match the predefined letter.
    \item Hard-2-back (NBH):The participant indicates when the current stimulus matches the stimulus that appeared two stages before.  
    \item Hard-under-Stress (NBS): The participant indicates when the current stimulus matches the stimulus that appeared two stages before (equivalent to NBH) but in the presence of additional noise (85dB) and white coat effect (presence of the experimenter in the room monitoring the participant performance).
    
\end{itemize} 

\textbf{The Operations Game task:} The activities in this task represent an assembly-line setting and test the fine motor skills of the participant. In this task, the participant is presented with a board in the form of a patient, and is asked to use tweezers to extract several objects from several slots, without touching the edges. This study too involves activities with a baseline, two difficulty levels and one stress condition. 

\begin{itemize}

    \item Baseline (DBB): The participant observes the board-game (operations) without touching the board and the objects. 
    \item Easy (DBE): The participant is asked to remove the five predefined objects (the easiest ones) within three minutes.
    \item Hard (DBH):The participant is asked to remove all twelve objects from the board within one minutes.   
    \item Hard-under-Stress (DBS): The participant is asked to remove all twelve objects from the board within one minute (equivalent to DBH) but in the presence of additional noise (85dB) and white coat effect (presence of the experimenter in the room monitoring the participant performance).
    
\end{itemize}

\textbf{The Webcall task:} The activities in this task represent the teleworking setting via tele-conferencing and ask for positive and negative memory recalls of the participant. In this task, the experimenter hosts an MS Teams call with the participant to hear them talking about their positive and negative memories or experiences. This task involves baseline, a positive and a negative condition lasting for about two minutes each. 
\begin{itemize}

    \item Baseline (WEB): The participant observes the monitor without talking. 
    \item Positive Condition (WEP): The participant recalls the happiest memory from their recent past.
    \item Negative Condition (WEN): The participant recalls a very negative / sad memory from their recent past.
\end{itemize} 

These tasks were executed in random order, and balanced across all participants. We refer to the data collected through this study and these tasks as Working-Environment-Context-Aware Dataset (WECARE-DB).

\begin{figure}[tb]
%\begin{subfigure}
  \centering
  % include first image
  \includegraphics[width=0.8\linewidth]{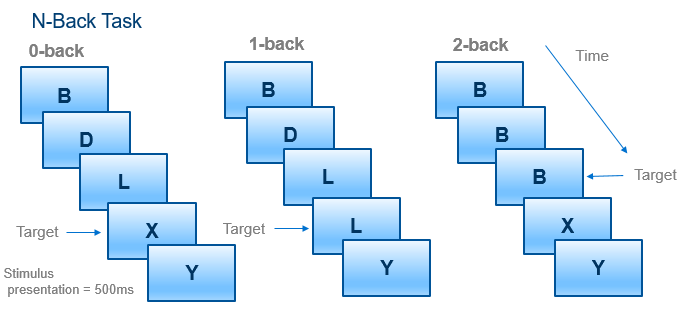}
  \caption{Example of N-back task with 0-back, 1-back and 2-back conditions}
  \label{fig:sub-first}
%\end{subfigure}
\end{figure}

\begin{figure}[tb]
  \centering
  % include second image
  \includegraphics[width=0.8\linewidth,  height= 5cm]{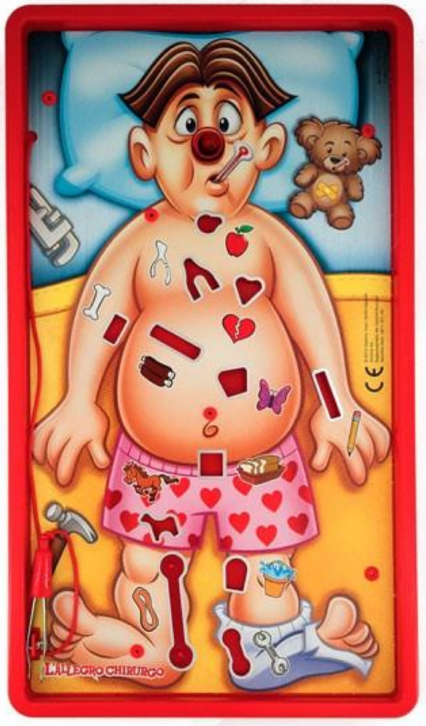}  
  \caption{The Operations Game task which instructs the participant to pick as many things out of the "patient" as possible without hitting the metal boundary. The nose will glow red and the board will vibrate if the participant makes an error.}
  \label{fig:sub-second}
\end{figure}

%%%%%%%%%%%%%%%%%%%%%%%%%%%%
% \begin{figure*}[hb]
%   \subfloat{
% 	\begin{minipage}[c][1\width]{
% 	   0.48\textwidth}
% 	   \centering
% 	   \includegraphics[width=0.9\textwidth, height = 5cm]{N-back-task.png}
% 	\end{minipage}}
%  \hfill 	
%   \subfloat{
% 	\begin{minipage}[c][1\width]{
% 	   0.48\textwidth}
% 	   \centering
% 	   \includegraphics[width=0.5\textwidth,  height = 5cm]{doctorgame.png}
% 	\end{minipage}}
% \caption{N-back Task and Doctor Game }
% \end{figure*}

\subsection{Questionnaires and participant self-report}

\noindent To compare the objective sensor data with subjective self-reported evaluations, a questionnaire called the \textbf{Self-Assessment Manikin (SAM)} was used. The Self-Assessment Manikin (SAM) is a picture-oriented questionnaire \cite{BradleyLang94}  developed to measure the valence/pleasure of the response (from positive to negative), perceived arousal (from high to low levels), and perceptions of dominance/control (from low to high levels) associated with a person's affective reaction to a wide variety of stimuli. The person is asked to provide only three simple judgments along each affective dimension (on a scale of 1 to 9) that best describes how they felt regarding the provided stimuli.
In our study, SAM was filled in after each baseline task and after each condition within a task. The questionnaire was introduced at the beginning of the experiment with example exercises. 

\begin{figure*}[h]
	\centering
	\includegraphics[width=17cm]{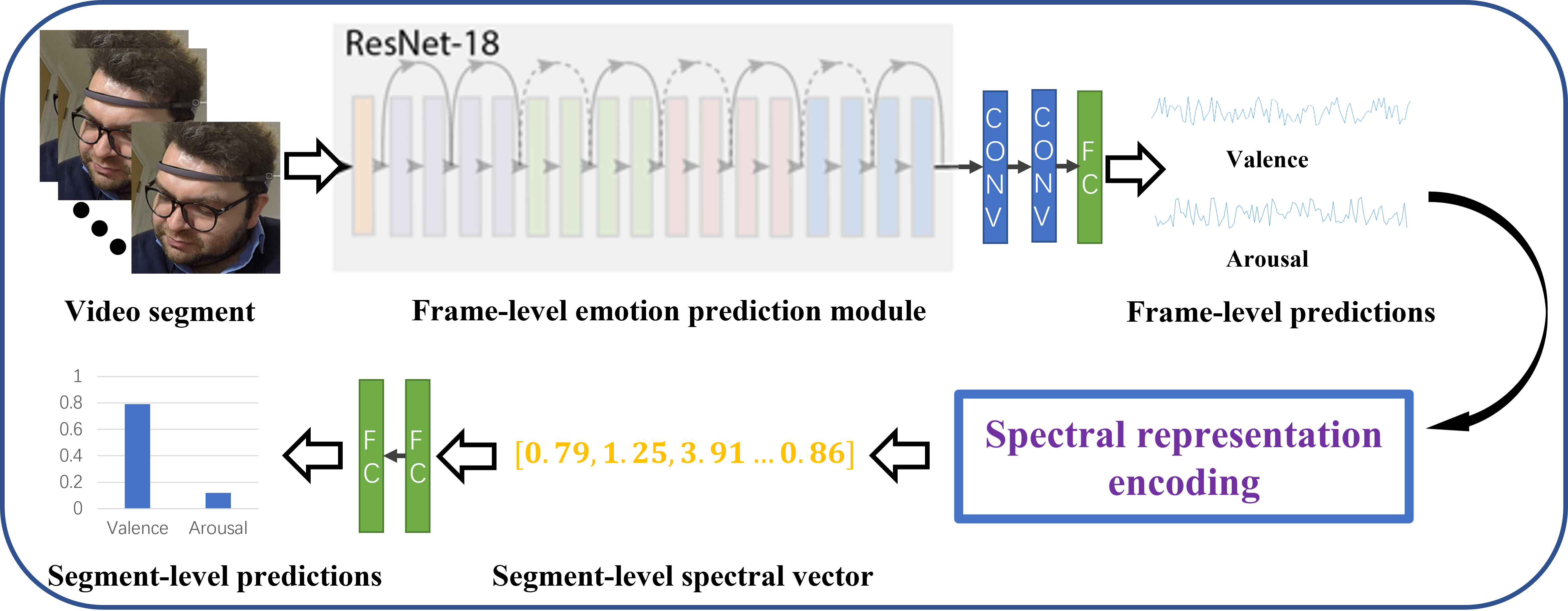}
	\caption{Illustration of the proposed S-ResNet framework.} 
    \label{fig:pipeline}
\end{figure*}

\section{Methodology}
\label{sec:method}

\noindent 
Firstly, to be able to use the self-reported SAM labels as \emph{ground truth}, we first map the collected values to valence and arousal dimensions. Specifically, we map the unhappy-happy dimension to valence and the calm-excited dimension to arousal. For both dimensions, we use $5$ as the threshold to map the corresponding values to negative ($<5$), neutral ($=5$) and positive ($<5$).

Secondly, to process the camera input, for each grabbed frame we apply face detection followed by facial landmark detection \cite{zhang2016joint}. The facial landmarks are utilized for face alignment \cite{baltrusaitis2018openface}. The aligned facial image is then fed to a $ResNet-18$ network \cite{he2016deep} with two additional convolution layers to provide deeper feature representation for valence/arousal prediction. This network is trained with a Mean Squared Error (MSE) loss to predict the valence and arousal values for each incoming facial image. The employed network is pre-trained using the AffectNet dataset \cite{mollahosseini2017affectnet}. This dataset contains large number of images from the Internet that were obtained by querying different search engines using emotion-related tags. 450,000 images in this dataset are manually annotated with valthe valence and arousal dimensions). 

Thirdly, we fine-tune this network using our collected WECARE dataset and we refer to this model as  $F-Res$. Finally, as the goal is to predict the affect of a user for a certain period in time, the spectral representation \cite{song2018human,song2020spectral} is utilised to summarize the  frame-level predictions for a certain period, which is then fed to another two-layer fully connected layer to generate the segment-level valence/arousal predictions. We refer to this model as $S-Res$. This pipeline is also illustrated in Fig. \ref{fig:pipeline}. 

The performance of all three models is evaluated using three metrics: Concordance Coefficient Correlation (CCC), Pearson Coefficient Correlation (PCC) and Root Mean Square Error (RMSE) as these are well known metrics utilised for automatic prediction of affect \cite{nicolaou2011continuous, mou2019your}.

%% Siyang modified
 
% \noindent For each collected frame, we first conduct face detection and alignment. The produced aligned facial image is then fed to a ResNet-18 network \cite{he2016deep} with two additional convolution layers to provide deeper facial representation for predicting valence/arousal intensities. This network is firstly pre-trained using the AffectNet dataset \cite{mollahosseini2017affectnet} which contains $450,000$ face images that are manually annotated with valence and arousal labels (labels range from -1 to 1 along both dimensions), where the Mean Squared Error (MSE) is employed as the loss function. After that, we fine-tuned this network using our collected WECARE data (we call this model as F-Res). Since the goal is to predict the dimensional affect of a certain period (multiple frames), the spectral representation \cite{song2018human,song2020spectral} is introduced to summarize frame-level predictions obtained by all frames of the target period, which is then fed to an MLP that contains two fully connected layers to generate the segment-level valence/arousal prediction (we call this model S-Res). This pipeline is our method is illustrated in Fig. \ref{fig:pipeline}. 

%%%%%%%%%%%%%%%%%%%%%%%%%%%%%%%%%%%

\begin{table}[tb]
\resizebox{0.49\textwidth}{!}{%
\begin{tabular}{@{}|l|l|l|l|l|l|l|@{}}
\toprule
\multirow{2}{*}{} & \multicolumn{3}{c|}{\textbf{Valence}}          & \multicolumn{3}{c|}{\textbf{Arousal}}          \\ \cmidrule(l){2-7} 
                  & ResNet-18 & F-Res & S-Res & ResNet-18 & F-Res & S-Res \\ \midrule
\textbf{RMSE}              & 0.38      & \textbf { 0.35}        & \textbf {0.35}        & 0.38      & 0.37        & \textbf {0.33}        \\ \midrule
\textbf{CCC}               & \textbf {0.56}      &\textbf { 0.56}        & 0.55        & 0.52      & 0.55        & \textbf {0.59 }       \\ \midrule
\textbf{PCC}               & 0.41      & 0.39        & \textbf {0.44}        & 0.38      & 0.43        & \textbf { 0.47}        \\ \bottomrule
\end{tabular}%
}\\
\caption{Valence and Arousal prediction results of the employed models in terms of RMSE, CCC and PCC. }
\label{tab:res}
\end{table}

%%%%%%%%%%%%%%%%%%%%%%%%%%%%%%%%%%%%%%%%%%%%%%%%%
\begin{figure*}[tb]
	\centering
	\includegraphics[width=16cm]{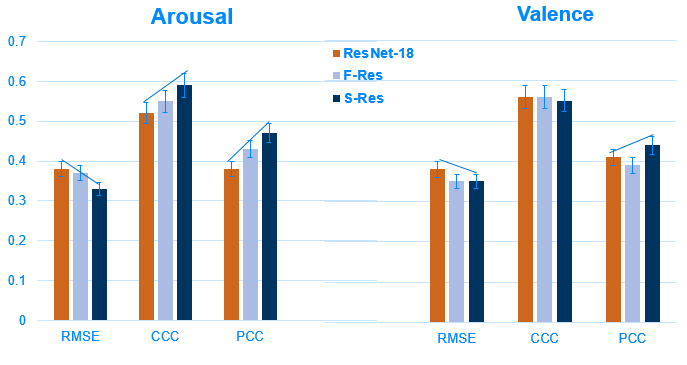}
	\caption{Illustration of Valence and Arousal prediction results of the employed models in terms of RMSE, CCC and PCC.} 
    \label{fig:model_res}
\end{figure*}

% \begin{figure}[t]
% 	\centering
% 	\includegraphics[width=9.2cm]{UCAM-VA_model_prediction.png}
% 	\caption{Illustration of Valence and Arousal prediction results of the employed models in terms of CCC, PCC and RMSE} 
%     \label{fig:pipeline}
% \end{figure}
%%%%%%%%%%%%%%%%%%%%%%%%%%%%%%%%%%%

% \usepackage{booktabs}
% \usepackage{multirow}
% \usepackage{graphicx}
\begin{table*}[]
\centering
\resizebox{\textwidth}{!}{%
\begin{tabular}{@{}cllllllllllll@{}}
\toprule
\multicolumn{2}{c}{\multirow{2}{*}{Tasks / Model}} & \multicolumn{4}{c}{\textbf{N-Back}} & \multicolumn{3}{c}{\textbf{Webcall*}} & \multicolumn{4}{c}{\textbf{Doctor Game}} \\ \cmidrule(l){3-6}\cmidrule(l){7-9}\cmidrule(l){10-13} 
\multicolumn{2}{c}{}                       & NBB  & NBE  & NBH  & NBS  & WEB  & WEP  & WEN  & DB   & DBE  & DBH  & DBS  \\ \cmidrule(r){1-2} \cmidrule(l){3-6}\cmidrule(l){7-9}\cmidrule(l){10-13}
\multirow{2}{*}{\textbf{ResNet-18}} & Mean & 0.33 & 0.33 & 0.25 & 0.16 & 0.25 & 0.41 & 0.08 & 0.08 & 0.16 & 0.33 & 0.25 \\
                                    & Std  & 0.49 & 0.49 & 0.45 & 0.38 & 0.45 & 0.51 & 0.28 & 0.28 & 0.38 & 0.49 & 0.45 \\ \cmidrule(l){2-13}
\multirow{2}{*}{\textbf{F-Res}}     & Mean & 0.41 & 0.49 & 0.33 & 0.25 & \textbf {0.31} & 0.51 & 0.16 & 0.33 & 0.25 & 0.33 & 0.33 \\
                                    & Std  & 0.52 & 0.55 & 0.49 & 0.45 & 0.51 & 0.52 & 0.38 & 0.49 & 0.45 & 0.49 & 0.49 \\ \cmidrule(l){2-13}
\multirow{2}{*}{\textbf{S-Res}}     & Mean & \textbf {0.51} & \textbf {0.51} & \textbf {0.58} & \textbf {0.67} & 0.16 & \textbf {0.58} & \textbf {0.25} & \textbf {0.56} & \textbf {0.75} & \textbf {0.68} & \textbf {0.56} \\
                                    & Std  & 0.52 & 0.52 & 0.51 & 0.49 & 0.38 & 0.51 & 0.42 & 0.51 & 0.44 & 0.47 & 0.51 \\ \cmidrule(l){2-13} 
\end{tabular}%
}
\caption{The average and standard deviation of the sign agreement results between model valence and arousal predictions and the self-reported SAM values. Tasks are divided with respect to the degree of task difficulty.}
\label{tab:tasks_div}
\end{table*}

%\begin{figure*}[tb]
%	\centering
%	\includegraphics[width=16cm]{Signagreement2.png}
%	\caption{Illustration of sign agreement results between model valence and arousal predictions and self-reported SAM questionnaire values with respect to the tasks performed. } 
%    \label{fig:sign_agreement}
%\end{figure*}

\section{Results and Discussion}
\label{sec:results_discussion}

\noindent Table \ref{tab:res} presents the valence and arousal predictions provided by three models: 1) ResNet-18 pre-trained by AffecNet \cite{mollahosseini2017affectnet} ($ResNet-18 model$); 2) ResNet-18 fine-tuned by our WECARE-DB ($F-Res model$); and 3) F-Res model with spectral representation-based segment-level prediction  ($S-Res model$). Looking at the table we observe that the model fine-tuned on WECARE-DB outperformed the ResNet-18 model pre-trained on AffecNet, indicating that the facial behaviours displayed in work-like environments are different compared to the in-the-wild Internet settings utilised in the AffectNet DB. Thus, it is necessary to acquire datasets and train models for recognising facial affect in work-like settings. Importantly, we found when using spectral representation \cite{song2018human,song2020spectral} to summarise segment-level information provides a large improvement suggesting that representing facial displays along time is crucial for predicting facial affect in work-like settings.

Table \ref{tab:tasks_div} presents the average and standard deviation of the sign agreement between the model predictions and self-reported SAM questionnaire. The values show that higher the agreement, the better the model performance. We observe that S-Res model that incorporates information from multiple frames performs best as compared to the other two models, thus supporting \textbf{RQ2}, that temporal information improves model performance. 

As future work, we will extend this study to a larger multi-site dataset that has been acquired at different European sites where the acquisition and study protocol used in this paper has been utilised to record participants performing work-like tasks. We will also investigate the information contained in other modalities such as psychological signals for predicting user affect when performing work-like tasks. The ultimate goal is to implement and use the trained models in real time and in real work settings to provide input to decision support systems to promote health and well-being of people during their working age in the context of the EU WorkingAge Project \cite{AlmeidaAAABBOCD20}.

\balance{}

\bibliographystyle{IEEEtran.bst}
\bibliography{main}\vspace{-5mm}

\begin{thebibliography}{10}
\providecommand{\url}[1]{#1}
\csname url@samestyle\endcsname
\providecommand{\newblock}{\relax}
\providecommand{\bibinfo}[2]{#2}
\providecommand{\BIBentrySTDinterwordspacing}{\spaceskip=0pt\relax}
\providecommand{\BIBentryALTinterwordstretchfactor}{4}
\providecommand{\BIBentryALTinterwordspacing}{\spaceskip=\fontdimen2\font plus
\BIBentryALTinterwordstretchfactor\fontdimen3\font minus
  \fontdimen4\font\relax}
\providecommand{\BIBforeignlanguage}[2]{{%
\expandafter\ifx\csname l@#1\endcsname\relax
\typeout{** WARNING: IEEEtran.bst: No hyphenation pattern has been}%
\typeout{** loaded for the language `#1'. Using the pattern for}%
\typeout{** the default language instead.}%
\else
\language=\csname l@#1\endcsname
\fi
#2}}
\providecommand{\BIBdecl}{\relax}
\BIBdecl

\bibitem{gunes2010automatic}
H.~Gunes and M.~Pantic, ``Automatic, dimensional and continuous emotion
  recognition,'' \emph{IJSE}, vol.~1, no.~1, pp. 68--99, 2010.

\bibitem{gunes2013categorical}
H.~Gunes and B.~Schuller, ``Categorical and dimensional affect analysis in
  continuous input: Current trends and future directions,'' \emph{Image and
  Vision Computing}, vol.~31, no.~2, pp. 120--136, 2013.

\bibitem{mohammadi2014pca}
M.~R. Mohammadi, E.~Fatemizadeh, and M.~H. Mahoor, ``Pca-based dictionary
  building for accurate facial expression recognition via sparse
  representation,'' \emph{Journal of Visual Communication and Image
  Representation}, vol.~25, no.~5, pp. 1082--1092, 2014.

\bibitem{Uddin2017Facial}
M.~Z. Uddin, M.~M. Hassan, A.~Almogren, A.~Alamri, M.~Alrubaian, and
  G.~Fortino, ``Facial expression recognition utilizing local direction-based
  robust features and deep belief network,'' \emph{IEEE Access}, vol.~5, pp.
  4525--4536, 2017.

\bibitem{ilyas2021deep}
C.~M.~A. Ilyas, M.~Rehm, K.~Nasrollahi, Y.~Madadi, T.~B. Moeslund, and
  V.~Seydi, ``Deep transfer learning in human--robot interaction for cognitive
  and physical rehabilitation purposes,'' \emph{Pattern Analysis and
  Applications}, pp. 1--25, 2021.

\bibitem{ilyas2018rehabilitation}
C.~M.~A. Ilyas, K.~Nasrollahi, M.~Rehm, and T.~B. Moeslund, ``Rehabilitation of
  traumatic brain injured patients: Patient mood analysis from multimodal
  video,'' in \emph{2018 25th IEEE ICIP}.\hskip 1em plus 0.5em minus
  0.4em\relax IEEE, 2018, pp. 2291--2295.

\bibitem{ilyas2020deepface}
C.~M.~A. Ilyas, A.~R. Nunes, M.~Rehm, K.~Nasrollahi, and T.~B. Moeslund, ``Deep
  emotion recognition through upper body movements and facial expressions,'' in
  \emph{16th VISAPP}.\hskip 1em plus 0.5em minus 0.4em\relax SCITEPRESS Digital
  Library, 2020, p. 229.

\bibitem{chen2018softmax}
L.~Chen, M.~Zhou, W.~Su, M.~Wu, J.~She, and K.~Hirota, ``Softmax regression
  based deep sparse autoencoder network for facial emotion recognition in
  human-robot interaction,'' \emph{Information Sciences}, vol. 428, pp. 49--61,
  2018.

\bibitem{mollahosseini2017affectnet}
A.~Mollahosseini, B.~Hasani, and M.~H. Mahoor, ``Affectnet: A database for
  facial expression, valence, and arousal computing in the wild,'' \emph{IEEE
  Transactions on Affective Computing}, vol.~10, no.~1, pp. 18--31, 2017.

\bibitem{calvaresi2017exploring}
D.~Calvaresi, D.~Cesarini, P.~Sernani, M.~Marinoni, A.~F. Dragoni, and
  A.~Sturm, ``Exploring the ambient assisted living domain: a systematic
  review,'' \emph{Journal of Ambient Intelligence and Humanized Computing},
  vol.~8, no.~2, pp. 239--257, 2017.

\bibitem{maskeliunas2019review}
R.~Maskeli{\=u}nas, R.~Dama{\v{s}}evi{\v{c}}ius, and S.~Segal, ``A review of
  internet of things technologies for ambient assisted living environments,''
  \emph{Future Internet}, vol.~11, no.~12, p. 259, 2019.

\bibitem{Reddy_2017_CVPR_Workshops}
B.~Reddy, Y.-H. Kim, S.~Yun, C.~Seo, and J.~Jang, ``Real-time driver drowsiness
  detection for embedded system using model compression of deep neural
  networks,'' in \emph{Proceedings of the IEEE CVPR Workshops}, July 2017.

\bibitem{Deng_Driver_2019}
W.~Deng and R.~Wu, ``Real-time driver-drowsiness detection system using facial
  features,'' \emph{IEEE Access}, vol.~7, pp. 118\,727--118\,738, 2019.

\bibitem{barsade2007does}
S.~G. Barsade and D.~E. Gibson, ``Why does affect matter in organizations?''
  \emph{Academy of management perspectives}, vol.~21, no.~1, pp. 36--59, 2007.

\bibitem{giorgi2021wearable}
A.~Giorgi, V.~Ronca, A.~Vozzi, N.~Sciaraffa, A.~Di~Florio, L.~Tamborra,
  I.~Simonetti, P.~Aric{\`o}, G.~Di~Flumeri, D.~Rossi \emph{et~al.}, ``Wearable
  technologies for mental workload, stress, and emotional state assessment
  during working-like tasks: a comparison with laboratory technologies,''
  \emph{Sensors}, vol.~21, no.~7, p. 2332, 2021.

\bibitem{borghini2020stress}
G.~Borghini, A.~Bandini, S.~Orlandi, G.~Di~Flumeri, P.~Aric{\`o}, N.~Sciaraffa,
  V.~Ronca, S.~Bonelli, M.~Ragosta, P.~Tomasello \emph{et~al.}, ``Stress
  assessment by combining neurophysiological signals and radio communications
  of air traffic controllers,'' in \emph{International Conference of the IEEE
  Engineering in Medicine \& Biology Society (EMBC)}.\hskip 1em plus 0.5em
  minus 0.4em\relax IEEE, 2020, pp. 851--854.

\bibitem{borghini2014measuring}
G.~Borghini, L.~Astolfi, G.~Vecchiato, D.~Mattia, and F.~Babiloni, ``Measuring
  neurophysiological signals in aircraft pilots and car drivers for the
  assessment of mental workload, fatigue and drowsiness,'' \emph{Neuroscience
  \& Biobehavioral Reviews}, vol.~44, pp. 58--75, 2014.

\bibitem{BradleyLang94}
M.~M. Bradley and P.~J. Lang, ``Measuring emotion: The self-assessment manikin
  and the semantic differential,'' \emph{Journal of Behavior Therapy and
  Experimental Psychiatry}, vol.~25, no.~1, pp. 49--59, 1994.

\bibitem{zhang2016joint}
K.~Zhang, Z.~Zhang, Z.~Li, and Y.~Qiao, ``Joint face detection and alignment
  using multitask cascaded convolutional networks,'' \emph{IEEE Signal
  Processing Letters}, vol.~23, no.~10, pp. 1499--1503, 2016.

\bibitem{baltrusaitis2018openface}
T.~Baltrusaitis, A.~Zadeh, Y.~C. Lim, and L.-P. Morency, ``Openface 2.0: Facial
  behavior analysis toolkit,'' in \emph{2018 13th IEEE international conference
  on automatic face \& gesture recognition (FG 2018)}.\hskip 1em plus 0.5em
  minus 0.4em\relax IEEE, 2018, pp. 59--66.

\bibitem{he2016deep}
K.~He, X.~Zhang, S.~Ren, and J.~Sun, ``Deep residual learning for image
  recognition,'' in \emph{Proceedings of the IEEE CVPR}, 2016, pp. 770--778.

\bibitem{song2018human}
S.~Song, L.~Shen, and M.~Valstar, ``Human behaviour-based automatic depression
  analysis using hand-crafted statistics and deep learned spectral features,''
  in \emph{2018 13th IEEE FG (2018)}.\hskip 1em plus 0.5em minus 0.4em\relax
  IEEE, 2018, pp. 158--165.

\bibitem{song2020spectral}
S.~Song, S.~Jaiswal, L.~Shen, and M.~Valstar, ``Spectral representation of
  behaviour primitives for depression analysis,'' \emph{IEEE Transactions on
  Affective Computing}, 2020.

\bibitem{nicolaou2011continuous}
M.~A. Nicolaou, H.~Gunes, and M.~Pantic, ``Continuous prediction of spontaneous
  affect from multiple cues and modalities in valence-arousal space,''
  \emph{IEEE Transactions on Affective Computing}, vol.~2, no.~2, pp. 92--105,
  2011.

\bibitem{mou2019your}
W.~Mou, H.~Gunes, and I.~Patras, ``Your fellows matter: Affect analysis across
  subjects in group videos,'' in \emph{2019 14th IEEE International Conference
  on Automatic Face \& Gesture Recognition (FG 2019)}.\hskip 1em plus 0.5em
  minus 0.4em\relax IEEE, 2019, pp. 1--5.

\bibitem{AlmeidaAAABBOCD20}
\BIBentryALTinterwordspacing
R.~M.~R. de~Almeida, A.~G. Aberturas, Y.~B. Aguado, M.~Atzori, A.~Barenghi,
  G.~Borghini, C.~A.~C. Ortega, S.~Comai, R.~L. Dur{\'{a}}n, M.~Fugini,
  H.~Gunes, B.~M. Lancis, G.~Pelosi, V.~Ronca, L.~Sbattella, R.~Tedesco, and
  T.~Xu, ``Decision support systems to promote health and well-being of people
  during their working age: The case of the workingage {EU} project,'' in
  \emph{Embedded Computer Systems: Architectures, Modeling, and Simulation -
  20th International Conference, {SAMOS} 2020, Samos, Greece, July 5-9, 2020,
  Proceedings}, 2020, pp. 336--347. [Online]. Available:
  \url{https://doi.org/10.1007/978-3-030-60939-9\_24}
\BIBentrySTDinterwordspacing

\end{thebibliography}

\end{document}